\documentclass[twoside,11pt,usehyper]{article}

\usepackage{blindtext}
\usepackage{setspace}
\usepackage{bold-extra}
\usepackage{amsmath}
\usepackage{eucal}
\usepackage{yfonts}
\usepackage{calligra}
\usepackage{wrapfig}
\usepackage{tcolorbox} 
\usepackage{lmodern}
\usepackage{titlesec}
\usepackage{microtype} 
\usepackage{lettrine} 
\usepackage[hyphens]{url}
\usepackage{breakurl}
\usepackage{xcolor}

\usepackage{jmlr2e}

\usepackage{lastpage}

\firstpageno{1}

\begin{document}

\onehalfspacing

\title{\huge{Function Alignment: \\ A New Theory of Mind and Intelligence} \\ \vspace{0.3em}\Large{ \normalfont \textit{Part I: Foundations}}}

\author{\name Gus G. Xia \\ 
       \addr Music X Lab, Machine Learning Department \\ Mohamed bin Zayed University of Artificial Intelligence
       }

\maketitle







\begin{tcolorbox}[width=0.72\textwidth, colback=white, colframe=black, boxrule=0.5pt, sharp corners, center] 
\raggedright 
\textit{
``The created universe carries Yin at the back and Yang in front, and through the union of the pervading principle, it reaches harmony.''
}
\vspace{0.5em} \\
\hfill \textit{---Laozi, Dao De Jing, Chapter 42} 
\end{tcolorbox}

\vspace{1em}
\noindent
\lettrine[lines=3, loversize=0]{$\mathcal{H}$}{} uman perception operates across multiple levels of abstraction simultaneously. For example, when we listen to music, we perceive raw acoustic signals at the most basic level, interpret musical scores at a higher level, and recognize even more abstract structures such as chords and forms. Representations at different levels function like distinct languages, each with its own semantics. Yet, these levels of representation influence one another, and we rely on such interactions to better understand and predict the world. Consider a music example again: a trained musician with theoretical knowledge can better anticipate upcoming low-level acoustic events, while an improviser attuned to the nuances of low-level music flow can make more informed decisions about which note to play next.
    

To model such entangled dynamics of hierarchical representations during perception, I propose \textbf{function alignment} as a new \textbf{theory of mind and intelligence} in this position paper. Note that this is not a technical paper in the engineering sense, but a methodological proposal. The theory is implementation-agnostic: in principle, it could be expressed through linear, nonlinear, probabilistic, or energy-based models—though from a modern AI perspective, a neural network implementation of function alignment is likely to offer the most natural blueprint for the human mind.

As shown in the graphical model in Figure~\ref{fig:illustration},  the $\boldsymbol{y}$-sequence represents the true dynamics of physical reality, the $\boldsymbol{x}$-sequence captures the low-level representation in the human mind, and the $\boldsymbol{z}$-sequence corresponds to a higher-level, more abstract representation. While additional layers may exist above $\boldsymbol{z}$, we limit our illustration to two levels for clarity. In this framework, the dynamics of $\boldsymbol{x}$ and $\boldsymbol{z}$ are defined as function aligned, characterized by the following three key properties:

\begin{figure}
    \centering
    \includegraphics[width=0.5\linewidth]{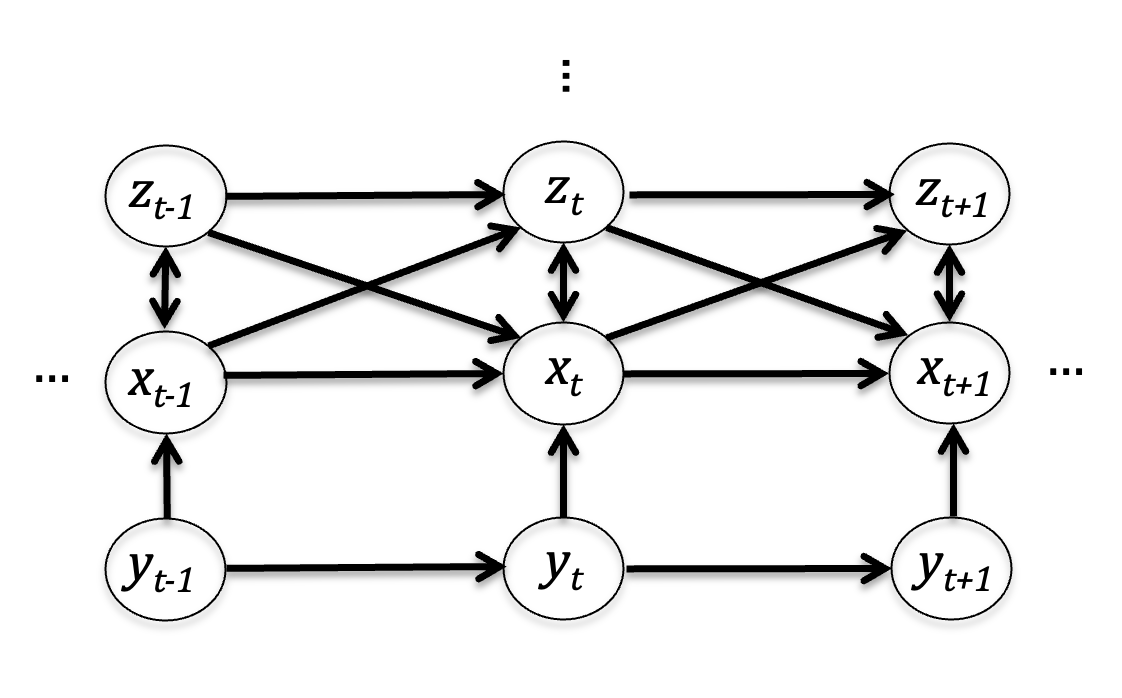}
    \caption{An illustration of function alignment: $\boldsymbol{x}$-$\boldsymbol{z}$ is function aligned, while $\boldsymbol{x}$-$\boldsymbol{y}$ is not.}
    \label{fig:illustration}
\end{figure}

\begin{enumerate}
    \item \textbf{Both $\boldsymbol{x}$ and $\boldsymbol{z}$ serve as functional descriptions of $\boldsymbol{y}$}---they encode different levels of abstraction of the same underlying reality.
    \item \textbf{Both $\boldsymbol{x}$ and $\boldsymbol{z}$ are auto-regressive processes, dynamically influencing each other’s predictions.} Unlike typical hierarchical models where higher layers passively summarize lower ones, here $\boldsymbol{x}$ and $\boldsymbol{z}$ actively ``listen'' to each other, forming a bidirectional alignment.
    \item \textbf{$\boldsymbol{x}$ and $\boldsymbol{z}$ are aligned in time,} where ``time'' refers not strictly to physical time but to a generalized logical sequence that governs inference and decision-making.
\end{enumerate}

The second property, represented by the diagonal cross-connections, is the most critical distinction between function alignment and conventional hierarchical time-series models. These cross-level, cross-time influences enable mutual adaptation rather than one-way abstraction. 

Notably, $\boldsymbol{x}$ and $\boldsymbol{y}$ are not function-aligned, as indicated by the absence of diagonal connections---$\boldsymbol{x}$ merely passively models the underlying reality $\boldsymbol{y}$ without influencing it. For instance, Newtonian mechanics provides a macro-scale description of physical reality based on underlying microscopic particle interactions, but the law $F = ma$ does not affect subatomic physics. Another example of a non-function-aligned hierarchical structure can be seen in programming: a Python program is interpreted into lower-level C-executable behavior, but the relationship is unidirectional. Such ``concealed'' hierarchical structures arise from a lack of function alignment. In contrast, the dynamics of the $\boldsymbol{x}$-sequence and the $\boldsymbol{z}$-sequence are deeply entangled and aligned---they are both perceptual representations, interconnected through neural mechanisms, and capable of shaping each other’s evolution.

\section*{Neural Hierarchical Representation and Symbolic Thinking}
\begin{wrapfigure}[]{r}{0.4\textwidth}
    \centering
    \includegraphics[width=0.9\linewidth]{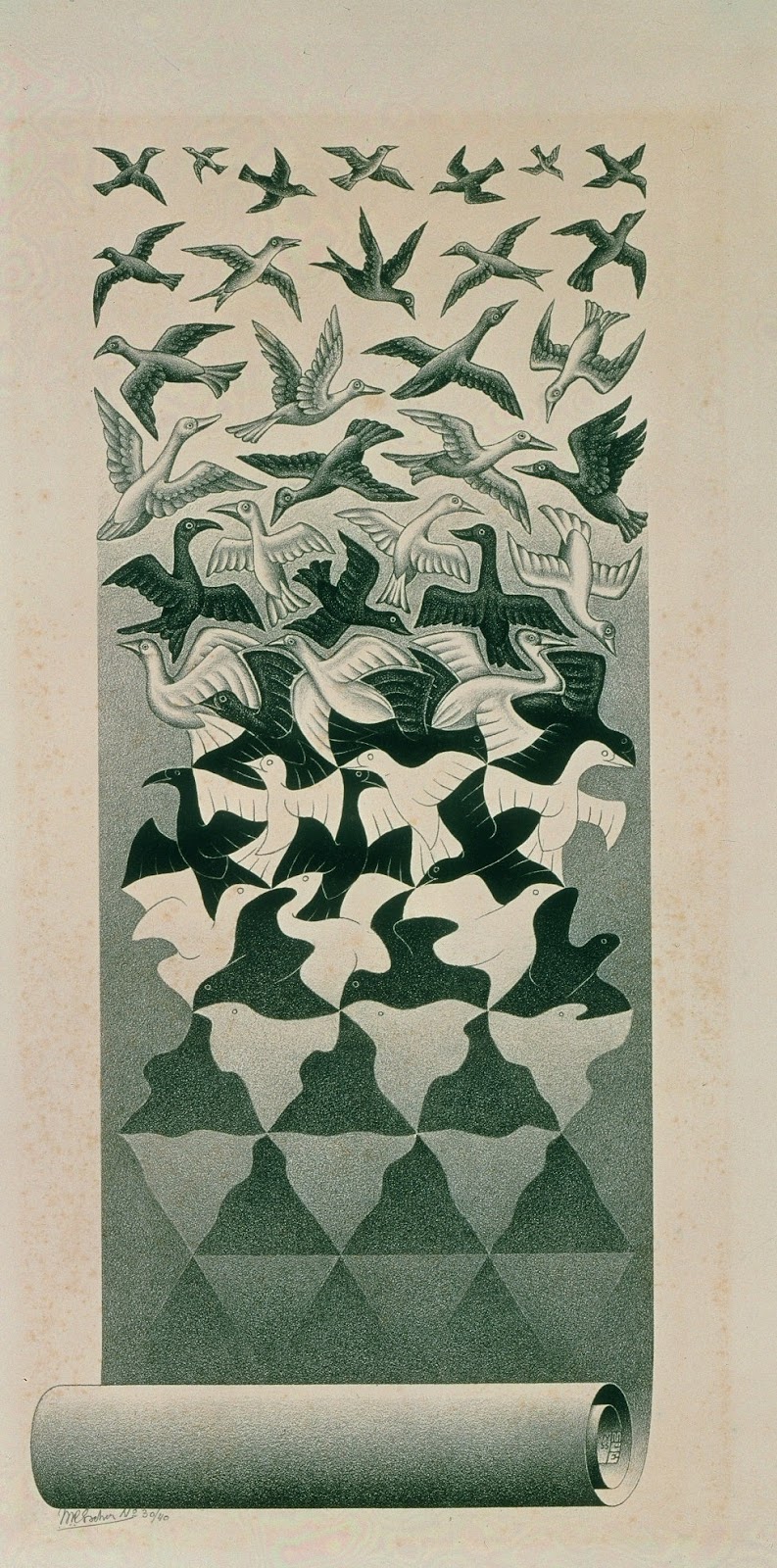}
    \caption{\textit{Liberation} by Escher.}
    \label{fig:rightfig}
\end{wrapfigure}

A plausible way to implement the interactions in Figure~\ref{fig:illustration} is through neural networks. As previously mentioned, we illustrate only two levels of representation, but in reality, the hierarchy could extend far beyond $\boldsymbol{z}$. The vertical bidirectional arrows between different levels of representation (e.g., $\boldsymbol{x}$ and $\boldsymbol{z}$) can be interpreted as an encoding-decoding mechanism, where higher-level representations are more abstract.

The encoding process inevitably leads to a loss of detail. At some sufficiently abstract level, representations reach what Hofstadter, in \textit{Gödel, Escher, Bach} \citep{geb}, calls the ``crystallization point'', where they transition into symbolic forms and their dynamics become rule-based. This hierarchical transition from abstraction to symbolism is conceptually similar to Escher’s artwork, where homogeneous, meaningless sub-symbolic forms at the bottom gradually evolve into distinct symbols with precise meanings at the top.

The focus of this section is not on how symbolic, rule-based processes emerge from sub-symbolic dynamics (this will be explored in the future part of this position paper). Instead, we examine how these two seemingly distinct processes---symbolic reasoning and neural computation---can function together as an organic whole, a relationship we refer to as function alignment.

As pointed out in the paper \textit{Thinking Like Transformers} \citep{weiss2021thinking}, rule-based programs can be ``compiled'' into a transformer. In other words, at the ``hardware'' level, the dynamics of some abstract representation still run on a neural system, while at the ``software'' level, the internal semantics can be a symbolic, rule-based program. 

Now, consider function alignment among the following three hierarchical levels of representation:
\begin{enumerate}
    \item The $\boldsymbol{x}$-sequence, representing subsymbolic dynamics.
    \item The $\boldsymbol{z}$-sequence, representing a neural-symbolic system, such as a Transformer that processes embeddings of symbols.
    \item The $\boldsymbol{z}'$-sequence, another neural-symbolic system, which shares the same symbol vocabulary as $\boldsymbol{z}$ but whose internal dynamics follow a compiled rule-based program.
\end{enumerate}

This framework provides a natural explanation for the \textbf{unity of intuition and rationality}---how humans can both follow rules and break them while simultaneously having an intuitive grasp of the underlying rationales. This flexibility lies in the ability to utilize different inference pathways at various levels of representation.

\section*{From Meaning to Explanation: The Bounds of Interpretation}

The function alignment framework certainly provides more insights---we can now offer precise and rigorous definitions for some deeply abstract and meta-level concepts central to mind and intelligence, including \textit{meaning}, \textit{interpretability}, \textit{analogy}, and \textit{rationality}.

\vspace{1em} \noindent
\textbf{Meaning}: What does a particular $\boldsymbol{y}$-sequence (some physical events) mean to us? Its meaning is simply \textit{the totality of hierarchical representations it triggers in the mind}. For instance, what might a short musical phrase mean to a trained musician? It could evoke subsymbolic-level emotional flow, symbolic-level representations such as melody, harmony, and form, and even highly abstract semantic layers, such as ``a lullaby my mother used to sing when I was a child,'' which can be verbalized in natural language.

Depending on context, different levels of representation may carry different weights of significance. For example, in the handwritten message, ``Top secret: the password to that old key safe is 9527,'' the core meaning clearly lies in the symbolic content---those four digits. But if this same sentence were found in a centuries-old letter, its value might shift toward its stylistic aspects, such as the aesthetic quality of its calligraphy---a subsymbolic representation.

Ideally, we would communicate by transferring the entirety of meaning---every level of representation that constitutes our internal feelings and thoughts---from one mind to another. But in practice, we cannot (at least not yet). Instead, we encode only selected layers into communicable media (such as text, speech, or music), and it falls to the receiver to \textit{interpret} the intended meaning based on their own internal function alignments.

\vspace{1em} \noindent
\textbf{Interpretability}, then, is the ability to express one representational layer (including $\boldsymbol{y}$, the lowest physical layer) in terms of another---whether from subsymbolic to symbolic, symbolic to subsymbolic, or from one symbolic layer to a more abstract one. For example, a parent might interpret their baby’s cry as ``I need food'' or ``I am cold''; a singer might interpret a symbolic score into a rich subsymbolic vocal performance; and in Eastern cultures, the phrase ``today’s moon is so beautiful'' might be interpreted as the abstract symbolic expression ``I love you.'' In the narrowest sense, interpretability is the ability to describe something using natural language---our shared symbolic representation layer $\boldsymbol{z}$.

Function alignment suggests that \textit{interpretability is intrinsically bounded}. Even when layers are aligned, they are not identical, and any interpretation inevitably sacrifices the unique dynamics of the original layer. For instance, interpreting a vocal performance through text will necessarily omit nuanced vocal expressions. Conversely, interpreting a written score through singing may miss explicit structural information such as harmonic progression.

Much has been debated \citep{fodor1988connectionism, walidblog} about whether artificial neural networks or human minds are ``interpretable.'' Within the function alignment framework, ``bounded interpretability'' clarifies that we can indeed interpret low-level neural outputs $\boldsymbol{x}$ using high-level symbolic representations $\boldsymbol{z}$, especially when agents share a common symbolic vocabulary. However, this power is limited: certain dynamics at the $\boldsymbol{x}$-level cannot be expressed within $\boldsymbol{z}$, regardless of alignment.

\vspace{1em} \noindent
\textbf{Analogy-making}, from this perspective, is a form of indirect interpretation. Instead of interpreting a target representation at one layer directly using a source representation from another layer, an analogy uses an alternative target that shares internal representations with the original. For example\footnote{Several examples here draw from the book \textit{Metaphors We Live By} (Lakoff \& Johnson, 2008)}:
\begin{itemize}
    \setlength{\itemsep}{1pt}
    \setlength{\parskip}{0pt}
    \item \textit{``Life is an adventure''} draws structure from purposeful, unpredictable journeys;
    \item \textit{``Argument is war''} relies on shared representations of attack, defense, and victory;
    \item \textit{``Your love is like moonlight''} evokes both an emotional tone and an elegant feeling.
\end{itemize}

In other words, analogy-making is interpretation via \textit{style transfer}: when two targets are juxtaposed, the human mind instinctively extracts their \textbf{shared source representation structure}. Sometimes, only the second (metaphorical) target appears, and the mind reconstructs the original target through context---such as in:
\begin{itemize}
    \item \textit{``He is the father of modern physics''}---where ``founder'' is omitted, yet readers infer not only his foundational role but also an emotional framing of responsibility, guidance, and care;
    \item \textit{``I cannot swallow that idea''}---where ``comprehend'' is replaced by a physical metaphor that conveys not only cognitive resistance but also a visceral sense of stress or discomfort---linking symbolic misunderstanding to subsymbolic embodiment.
\end{itemize}

This ``shared source'' view of analogy aligns closely with the notion of “conceptual skeletons” proposed in \textit{Gödel, Escher, Bach} \citep{geb} and in \textit{Surfaces and Essences} \citep{sae}. And the function alignment framework extends this view further: the shared representation need not reside solely at an abstract symbolic level. It may span multiple representational layers---embodied, subsymbolic, symbolic---and the more layers involved in alignment, the greater the analogy’s explanatory and expressive power.

Many Zen koans and classical poetry draw on natural phenomena not to illustrate abstract notions but to evoke \textit{subsymbolic felt states} across layers. For example, \textit{``the body is like the Bodhi tree; the mind is like a bright mirror''} encodes not just visual and symbolic meaning, but also an introspective experiential resonance. So, too do the aforementioned analogies of ``moonlight'' and ``swallow.'' In these cases, analogies become a conduit for multi-layered profound experience.

Analogies are not only powerful but also ubiquitous and sometimes hard to be noticed. Consider the simple act of pointing to an object and saying its name---``this is a chair''---is actually a form of analogy. The visual form of the chair is $\boldsymbol{y}$, the acoustic form of the chair is $\boldsymbol{y'}$, and through function alignment, both $\boldsymbol{y}$ and $\boldsymbol{y'}$ will trigger the same symbolic-level representation $\boldsymbol{z} = \text{``chair''}$. Despite its effectiveness, analogy is bounded in the same way as interpretability: we rely on partially aligned internal representations to make sense of each other’s outputs. When alignment breaks down, interpretation collapses into nonsense.

\vspace{1em} \noindent
\textbf{Explanation}: We are now ready to define explanation within the function alignment framework. To explain any target is to reveal the causal representations—those pointing to the target—within the function alignment process, through interpretation and analogy. This aligns with the tradition of causal modeling, where an explanation identifies structural dependencies, often formalized as directed acyclic graphs (DAGs). Indeed, the function alignment framework itself can be viewed as a dynamic causal model, where vertical bidirectional arrows function as real-time feedback, without violating local acyclicity.

However, our view departs from traditional causal modeling in a subtle but crucial way: we regard explanation not just as ``passive description", but as ``active prescription." Most conventional explanations---whether graphs or language---are z-level \textit{communicable representations}, which are ``symbolic shadows" of an underlying structure. To truly understand an explanation is not just to parse its symbolic content, but to align it with subsymbolic flows. E.g., one may study tonal music theory through graphs or language, but its real explanatory power emerges only when it is felt and aligned with auditory and embodied perceptions.

In traditional causal theory, a flawed model can simply be updated. However, under function alignment, a misaligned mental model may lead to explanatory closure and even interfere with the system it claims to describe. Consider a musician clinging to diatonic theory while listening to a microtonal piece: the sounds are heard, but perception collapses. Worse, if the musician is performing and imposes this frame on others, the shared experience becomes forcibly re-centered around the wrong model.

In this sense, explanation becomes a double-edged tool: \textit{when aligned, it reveals; when misaligned, it blinds.} The key question is no longer ``Is this explanation true?" but rather:
``Is the explanation well-aligned? Is it timely, adaptive, and capable of shaping the very process it seeks to explain?" Truth becomes not a static match, but a dynamic resonance—between model and flow, between language and life.

\vspace{1em} \noindent
\textbf{Rationality}, therefore, is the capacity to explain using logical or symbolic language. As we have argued, such interpretability is bounded---symbolic reasoning typically occurs at higher-level representational layers, while actual behavior and perception are rooted in lower layers. This view provides a structural and representational grounding for Herbert Simon’s theory of \textit{bounded rationality} \citep{simon1990bounded}. Simon proposed that humans are rational within limits---capable of reasoning, but constrained by cognitive resources. We now have a structural mechanism underlying this description: the limited cognitive capacity for rationality arises from the bounded interpretability inherent in function alignment. When an agent explains a satisficing action using a logic-based framework, it performs an interpretive operation that is, by definition, partial and bounded. People do, in fact, optimize across all levels of representation---including embodied feelings and subsymbolic dynamics---which, in Simon’s terms, is to ``satisfice.'' However, such optimization cannot be fully explained at the rational, symbolic level.

In contrast, \textit{complete explanation} is possible only within a pure formal system. In such systems, all reasoning and representation occur within a single symbolic layer. There is no information loss about alignment, because no cross-level interpretation is required; explanation collapses into logical operations, and all causes and effects are expressed in the same formal vocabulary. Thus, the ideal of perfect rationality coincides with perfect alignment---achievable only in formal systems.

\section*{Agent-Based Intelligence and Isomorphic Alignment}

We now formalize function alignment within a mathematical framework. Again, though many layers of representation may exist, here we only formulate two layers for simplicity. Without losing generality, the function alignment between $\boldsymbol{x}$ and $\boldsymbol{z}$ can be modeled as a coupled system:
\begin{align}
x_t &= f(\boldsymbol{x}_{<t}, \boldsymbol{z}_{<t}, z_t) \\
z_t &= g(\boldsymbol{z}_{<t}, \boldsymbol{x}_{<t}, x_t)
\end{align}
Here, $f$ and $g$ can potentially be any form of function representing the forward dynamics. The entanglement arises because each update depends on the other's past and current state. Specifically, if we want to distinguish between the cross-time mapping and the coupling effect at each time point, the system can be alternatively formulated as:
\begin{align}
    x_t &= f(\boldsymbol{x}_{<t}, \boldsymbol{z}_{<t}) \\
    z_t &= g(\boldsymbol{z}_{<t}, \boldsymbol{x}_{<t})
\end{align}
\vspace{-3em}
\begin{align}
\text{subject to} \quad E(x_t, z_t) < \delta
\end{align}
where $E$ can be regarded as a soft energy function enforcing structural compatibility and semantic isomorphism, rather than perfect reconstruction.
In practice, abstract representations $z$ almost always lose some information of $x$, so the alignment is inherently asymmetric. That is, while $x$ may fully determine $z$, the reverse is not generally true. Nonetheless, by maintaining $E(x_t, z_t)$ within a bounded tolerance $\delta$, we preserve semantic coherence across layers. We can therefore define a loss term:
\begin{equation}
\mathcal{L}_{\text{FA}} = 
\alpha_1 \cdot \text{dist}(x_t, f(\boldsymbol{x}_{<t}, \boldsymbol{z}_{<t})) + 
\alpha_2 \cdot \text{dist}(z_t, g(\boldsymbol{z}_{<t}, \boldsymbol{x}_{<t})) + 
\beta \cdot E(x_t, z_t)
\end{equation}

\vspace{1em} \noindent
\textbf{Isomorphic alignment among agents}:  
We now extend the function alignment framework to multi-agent scenarios. Consider two individual autoregressive processes (agents), each evolving with the same functional form:
\begin{align}
    \boldsymbol{x}_t &= h_{\theta_x}(\boldsymbol{x}_{<t}) \\
    \boldsymbol{z}_t &= h_{\theta_z}(\boldsymbol{z}_{<t})
\end{align}
Here, $h_{\theta_x}$ and $h_{\theta_z}$ represent the internal dynamics of each agent. If these processes describe different aspects of a shared underlying phenomenon—such as performance-level and harmony-level dynamics in music—we may apply a post-hoc function alignment to couple them into a unified system. This converts the initially independent agents into a jointly evolving structure, as formalized earlier in Equations (1–5). 

The notion of \textit{isomorphic alignment} refers to the condition where the aligned system can be written as:

\begin{align}
    \boldsymbol{w}_t &= [\boldsymbol{x}_t, \boldsymbol{z}_t] \\
    \boldsymbol{w}_t &= h_{\theta_w}(\boldsymbol{w}_{<t})
\end{align}

where $[\boldsymbol{x}_t, \boldsymbol{z}_t]$ denotes the concatenation of the two states at time $t$, and $\theta_w$ represents the unified dynamics after alignment.

Isomorphic alignment occurs when the unified dynamics $h_{\theta_w}$ preserves the same functional form as the original agents $h_{\theta_x}$ and $h_{\theta_z}$—differing only in parameterization. In this sense, the aligned system is structurally isomorphic to its components. This mirrors the idea of conjugacy in Bayesian inference, where the prior and posterior belong to the same distributional family. However, in this case, the isomorphism results not from within-model temporal updating, but from structural alignment across multiple generative processes.

\vspace{1em} \noindent
\textbf{Linear case}: To better illustrate the idea of isomorphic alignment, let us consider a simple linear case. If we regard Figure~\ref{fig:illustration} as a first-order linear dynamical system, horizontal and diagonal arrows across time steps represent transition matrices, while vertical arrows within a single time step reflect instantaneous coupling, captured via covariance matrices. The aligned systems can be formulated as:
\begin{align}
    x_t &= Ax_{t-1}+Cz_t \\
    z_t &= Bz_{t-1}+Dx_t 
\end{align}
We use \textit{degrees of freedom} (DoF) to characterize the structure. Suppose that before alignment, $\boldsymbol{x} = \{{x}_1, {x}_2, \dots\}, {x}_i \in \mathbb{R}^n$ and $\boldsymbol{z} = \{{z}_1, {z}_2, \dots\}, {z}_i \in \mathbb{R}^m$ are two independent linear dynamical processes, each governed by:
\begin{align}
    x_t &= A'x_{t-1} \\
    z_t &= B'z_{t-1}
\end{align}
Their degrees of freedom are:
\begin{itemize}
    \item $\texttt{DOF}_{\boldsymbol{x}}$: $n^2$ for transition + $n(n+1)/2$ for initial covariance
    \item $\texttt{DOF}_{\boldsymbol{z}}$: $m^2$ for transition + $m(m+1)/2$ for initial covariance
\end{itemize}
Now consider a combined process ${w}_t = [{x}_t, {z}_t] \in \mathbb{R}^{n+m}$ modeled as a first-order linear system, whose degrees of freedom are:

\[
\texttt{DOF}_{\boldsymbol{w}} = (n + m)^2 + \frac{(n + m)(n + m + 1)}{2}
\]

Function alignment imposes three types of structural connections:  
(1) horizontal autoregression, (2) vertical intra-step coupling, and (3) diagonal cross-step feedback.

\begin{itemize}
    \setlength{\itemsep}{1pt}
    \item Horizontal transitions: $\boldsymbol{x}_t \rightarrow \boldsymbol{x}_{t+1}$ and $\boldsymbol{z}_t \rightarrow \boldsymbol{z}_{t+1}$, contributing $\texttt{DOF}_{\boldsymbol{x}}+\texttt{DOF}_{\boldsymbol{z}}$
    \item Vertical coupling (within-time): $nm$ DoF
    \item Diagonal alignment (cross-time interaction): additional $2nm$ DoF
\end{itemize}

Putting all these together, we observe that the function-aligned system recovers the full degrees of freedom:
\[
\texttt{Total function-aligned DoF} = \texttt{DOF}_{\boldsymbol{x}} + \texttt{DOF}_{\boldsymbol{z}} + 3nm = \texttt{DOF}_{\boldsymbol{w}}
\]
This matches exactly the DoF of the unified system $\boldsymbol{w}$, showing that \textbf{full function alignment renders two interacting representational processes structurally isomorphic to a single unified linear dynamical agent}. If any of the alignment arrows—horizontal, vertical, or diagonal—is missing, the system loses structural completeness and can no longer be regarded as fully integrated. This confirms that full function alignment achieves structural isomorphism. \hfill Q.E.D.

\vspace{1em} \noindent
\textbf{Nonlinear and long-dependency generalization}:  
In nonlinear or higher-order systems, degrees of freedom may no longer be analytically tractable. In such cases, we may resort to alternative structure criteria, such as symmetry, gradient continuity, or variational closure.

Moreover, when alignment extends across longer time horizons or spans multiple representational agents, \textit{strict} isomorphic alignment---i.e., full connectivity where everything connects to everything---may not be necessary and achievable. For instance, in the brain, inter-hemispheric links are often far sparser than intra-regional connections. What matters is not universal connectivity, but the presence of functional bridges across abstraction layers, enabled directly or indirectly via the “diagonal arrows” in our framework. We refer to such generalized cases as \textit{relaxed isomorphic alignment}.

Under this broader view, the principle remains: alignment must preserve the system’s capacity for dynamic co-adaptation across time and abstraction. When cross-layer pathways, especially diagonal arrows, are broken, the system fragments, introducing adaptation bottlenecks or isolating representational islands. Alignment integrity thus becomes a necessary condition for unified learning and intelligent behavior.

\vspace{1em} \noindent
\textbf{Implication}:  
This structure demonstrates that function alignment is not only a perceptual architecture but also a structural condition for agent unification. Different components of the mind—or the brain—may specialize in distinct representational dynamics, yet through function alignment, they can operate as a unified cognitive agent. More generally, simple intelligent agents at the micro-level can be recursively aligned to form macro-level agents, while preserving relaxed isomorphic structure. This recursive principle offers a scalable blueprint for constructing life-like intelligence, in both artificial and biological systems.

\section*{\textls[-10]{Beyond Modeling: Insights for Psychology, Philosophy, and Zen}}

Function alignment offers not only a computational theory of mind but also a lens through which we can deepen our understanding of ourselves. Across psychology, philosophy, and contemplative traditions like Zen, we encounter a core duality: the mind has two systems---one intuitive, one analytical. The challenge has always been not to choose one over the other, but to integrate them in a way that leads to wisdom and harmony.

In cognitive science, the duality is known as \textbf{System 1} and \textbf{System 2} \citep{kahneman2011thinking}; in AI, \textbf{Mode 1} and \textbf{Mode 2} (e.g., Hierarchical JEPA \citep{jepa}); in traditional philosophies, \textbf{Yin} (or feminine) and \textbf{Yang} (or masculine); in \textit{Zen and the Art of Motorcycle Maintenance} \citep{zenmotor}, \textbf{romantic} and \textbf{classical} understanding. A harmonic integration of these two modes is not merely an intellectual task, but a living art and experience.

\vspace{1em} \noindent
\textbf{Split brains and minds}: Psychological studies, especially on split-brain patients, offer striking evidence of function alignment and agent-based intelligence, even at the physical brain level. In one well-known setup \citep{charge}: when light is shown to the left visual field, the right hemisphere perceives it, but the left hemisphere (responsible for language) cannot report it via natural language at the $\boldsymbol{z}$ level. Still, the subject, at the behavior $\boldsymbol{x}$-level can press the button to indicate ``seeing the light.'' This suggests that each hemisphere can function as an agent, but only the left brain has access to symbolic language expression. Without aligned input from perception, reasoning is blind.

\vspace{1em} \noindent
\textbf{Koan of ``mirror becomes the mask''}: If split-brain patients show what happens when symbolic processing is disconnected, our daily struggle is often the opposite: an \textit{over-identification with the symbolic layer}, just as a performer who overemphasizes music theory may lose touch with the underlying flow of music itself. All minds are function-aligned but to varying degrees. A sharp rational mind is smart, but only a deeply aligned mind is wise. My favorite story about this second kind of misalignment comes from Sadhguru:

\begin{quote}
    \centering
    \textit{A man, having promised to quit drinking, again returned home late and drunk. On the way, he scratched his face on a branch. Not wanting his wife to know, he quietly applied bandages in the bathroom and sneaked into bed without a sound. \\ \vspace{5pt}
    The next morning, his wife slapped him: ``You drank again!'' He was shocked: ``How did you know?'' She pointed at the mirror. ``The bandages were all over it!''}
\end{quote}

This is indeed a great metaphor of misalignment: $\boldsymbol{z}$ is the mirror of $\boldsymbol{x}$, and when aligned, $\boldsymbol{z}$ should reflect and serve $\boldsymbol{x}$. But when we are not conscious enough, misalignment arises, and $\boldsymbol{z}$ becomes ego and a distortion we confuse with truth. Despite smartness, we may act for value detached from experience, argue logic without feeling, and become minds that speak without seeing. 

\vspace{1em} \noindent
\textbf{Towards integration}: The practice of Zen is, in essence, training in experiencing the truth and a deep function alignment. Enlightenment is not knowing more intellectually but \textbf{reconnecting to the raw flow of perception $\boldsymbol{x}$} that is as close to $\boldsymbol{y}$ as possible, followed by a non-egoic reintroduction of symbolic framing and deep function alignment. The Zen master’s sudden shout, the nonsensical koan, and the silent meditation practice of ``just sit''  are designed not to teach knowledge \textit{about} truth, but to interrupt overactive $\boldsymbol{z}$-level symbolic thinking and restore access to the unfiltered experience.

As a well-known Zen analogy goes:

\begin{quote}
    \centering
    \textit{First, mountains are mountains. \\
    Then, mountains are not mountains. \\
    Finally, mountains are once again mountains.}
\end{quote}

These three stages beautifully trace the arc of function alignment:
\begin{itemize}
    \setlength{\itemsep}{1pt}
    \setlength{\parskip}{0pt}
    \item The first stage reflects a symbolic-dominant misalignment---where socially conditioned values obscure unfiltered experience.
    \item The second stage marks disorientation from symbols, allowing one to contact direct perceptual truth.
    \item The third stage represents deep re-integration, where symbols are rebuilt---fresh, aligned, and transparent.
\end{itemize}

In brief, this is not a rejection of rationality but a transcendental path where symbols are no longer a substitute for truth but its humble servant. It is the arc of function alignment in full: from confusion, to liberation, to a deep ease of being---a mind no longer at war with itself but aligned in harmony.

\section*{Conclusion and Outlook}
In conclusion, this paper proposes \textit{function alignment} as a theory of mind that is not only intuitively compelling, but structurally grounded. Unlike many existing accounts of cognition that rely on pre-theoretical concepts or loosely specified metaphors, this framework makes explicit how meaning, interpretation, and analogy emerge from concrete relationships among representational layers. Each concept introduced, whether symbolic reasoning or feeling-level resonance, corresponds to a definable pattern of interaction within the alignment model. In this sense, function alignment forms a coherent representational language, capable not only of modeling minds, but of serving as a blueprint for building them.

Moreover, this theory does something unique: \textit{it explains explanation, and it gives meaning to the very concept of meaning}. It shows why interpretation is inherently bounded---meaning is layered, and it must be aligned to be understood. This perspective offers a unified theoretical grounding for many fragments of mind science, such as bounded rationality, symbol grounding, and analogy-making. Once treated as isolated phenomena and concepts, they now emerge as structural consequences of representational dynamics. 

Furthermore, function alignment bridges domains too often kept apart. It is not built upon any philosophy or belief system. Rather, philosophies, psychologies, and even contemplative systems like Zen may find themselves reconstructible within it. If symbolic thought is to serve experience rather than obscure it, then we need not only more knowledge, but better alignment. Function alignment offers a shared foundation where logic and perception, explanation and intuition, can meet---not in conflict, but in coherence and harmony.

Finally, this first part has focused on the foundational aspects of the function alignment framework. It leaves several critical aspects for future development: the nature of action, interaction with environments and other agents, the emergence of symbolic language, and the question of how such alignment mechanisms might be realized via AI systems. As a foundational entry, this work sets the stage for these exciting developments to come in a larger program.

\acks{I would like to thank Roger Dannenberg, Yann LeCun, He He, and Maigo Wang for their insightful discussions on hierarchical modeling. Also, I would like to thank Chao Shi and Rongfeng Li for the discussion on the mathematical formulation of function alignment.
I am grateful to Liwei Lin, Junyan Jiang, Yuxuan Wu, Ziyu Wang, and Daniel Chin for their contributions to the initial development of function alignment, as well as their support with pilot studies, experiments, and paper formatting.}



\newpage









\vskip 0.2in
\bibliography{ref}

\end{document}